\documentclass[10pt,twocolumn,letterpaper]{article}

\usepackage{wacv}

\usepackage{graphicx}
\usepackage{booktabs}
\usepackage{multirow}
\usepackage{array}
\usepackage{tabularx}
\usepackage{threeparttable}
\usepackage{placeins}

\usepackage{amsmath}
\usepackage{amssymb}
\usepackage{bbm}

\usepackage{algorithm}
\usepackage{algorithmic}

\usepackage{xcolor}
\usepackage[normalem]{ulem}        
\newcommand{\rev}[1]{#1}           

\usepackage[breaklinks=true,bookmarks=false]{hyperref}

\def\confName{WACV}
\def\confYear{2027}

\title{VTRL-Med: Progressive Visual Refinement with
Topology-Guided Regularization for Medical Image Clustering}

\author{
Guangyu Meng$^{1}$ \qquad
Pengfei Gu$^{2}$\thanks{Corresponding authors. This research was supported in part by NSF Grants CCF-2444309, CCF-2523787, and IIS-2403438, and AHA award 26AIREA1574568.} \qquad
Peixian Liang$^{1}$ \\[0.3em]
John P. Lalor$^{1,3}$ \qquad
Erin Wolf Chambers$^{1}$\footnotemark[1] \qquad
Danny Z. Chen$^{1}$\footnotemark[1]\\[0.5em]
$^{1}$Dept.\ of Computer Science and Engineering, University of Notre Dame\\
$^{2}$Dept.\ of Computer Science, The University of Texas Rio Grande Valley \\ \quad
$^{3}$Dept.\ of IT, Analytics, and Operations, University of Notre Dame\\[0.3em]
{\tt\small \{gmeng, pliang, john.lalor, echambe2, dchen\}@nd.edu, \quad pengfei.gu01@utrgv.edu}
}
\begin{document}
\maketitle

\begin{abstract}
Deep clustering offers a promising approach for analyzing large unlabeled medical image datasets, yet prevailing methods still face two limitations: they may neglect class-agnostic structural priors that capture object boundaries and regions, and they do not explicitly encode topological patterns (e.g., connectivity and holes). 
These sources of information play distinct and complementary roles: visual-structural features provide the primary discriminative representation for clustering, while topological features provide additional global geometric cues for residual visual ambiguities. 
We introduce VTRL-Med, a progressive visual-structural learning framework with topology-guided regularization for medical image clustering. VTRL-Med first adapts the Segment Anything Model (SAM) to extract image features that retain structural cues inherited from mask-supervised pretraining. It then generates reliable pseudo-labels through a two-step filtering process that combines prototype-based selection with k-nearest neighbor consistency validation. 
Finally, it progressively refines the visual branch while
an auxiliary topology branch, constructed from persistent homology (PH) descriptors, provides additional geometric regularization during training. Across 13 medical image benchmarks, VTRL-Med achieves state-of-the-art clustering performance, improving average accuracy by 4.48\% over existing methods. On three representative datasets, topology-guided
regularization further improves accuracy by 1.3--4.3\% over visual
refinement alone.
\end{abstract}

\section{Introduction}
\label{sec:intro}

Medical image analysis is critical for disease diagnosis and biomarker discovery~\cite{litjens2017survey}, but supervised deep learning methods face a major bottleneck: expert annotations are costly, time-consuming, and subject to inter-observer and intra-observer variability~\cite{irvin2019chexpert, rajpurkar2017chexnet,meng2026topocl}. For example, pixel-level labeling in pathology demands extensive expert labeling effort~\cite{campanella2019clinical}, yet many clinically relevant phenotypes cannot be defined a priori for annotation~\cite{seebock2018unsupervised}. Consequently, massive unlabeled datasets are underutilized while small labeled datasets limit DL model robustness. Label-free clustering can support critical clinical needs beyond annotation, including discovery of novel disease subtypes~\cite{wu2017unsupervised,muhammad2019unsupervised}, cohort stratification for treatment response prediction~\cite{wu2017unsupervised}, and quality control in large-scale screening programs~\cite{litjens2017survey}.

\begin{figure}[t]
\centering
\includegraphics[width=\columnwidth,trim={0cm 2cm 0cm 0cm},clip]{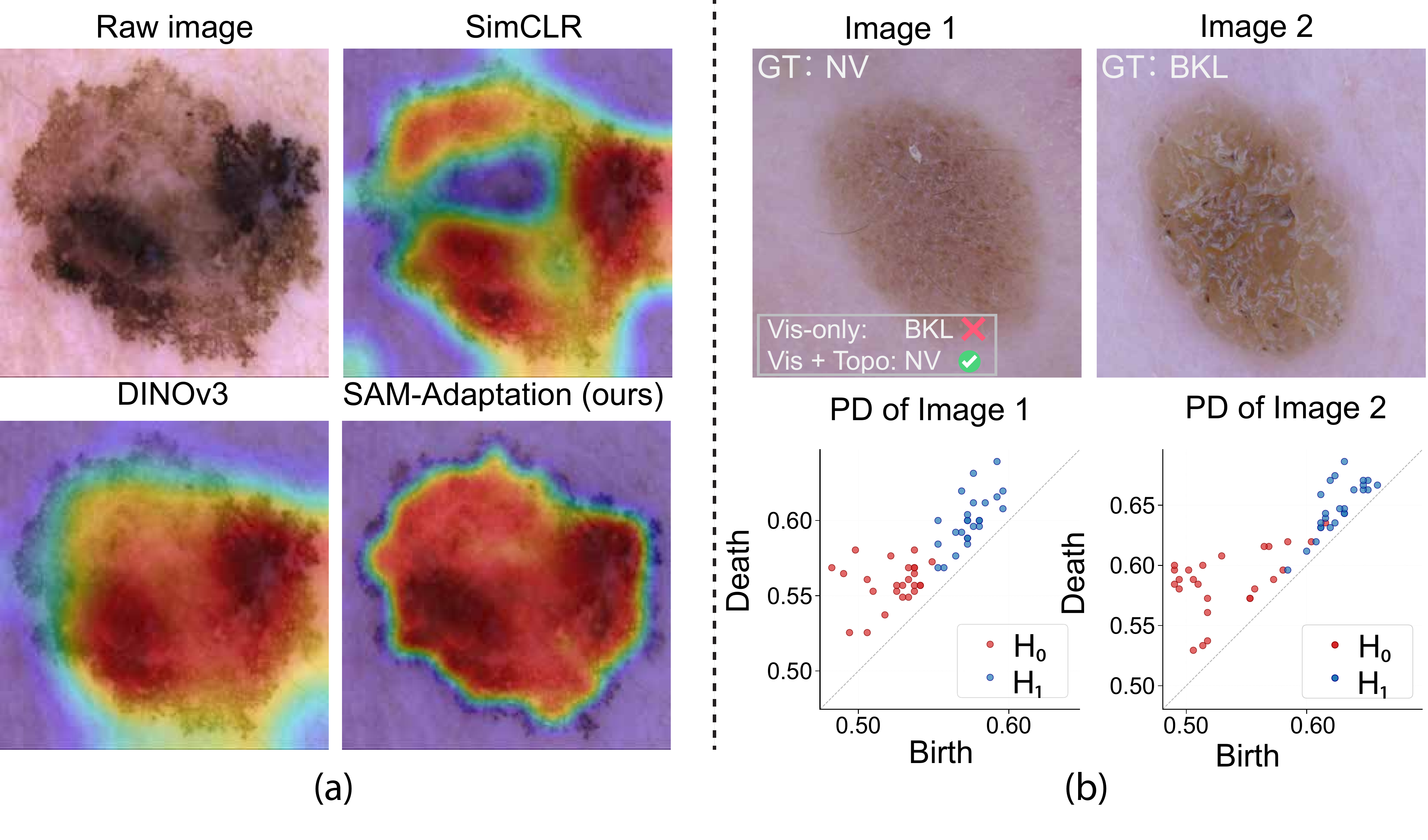}
\caption{
Motivation for VTRL-Med.
\textbf{(a)} Grad-CAM visualizations~\cite{selvaraju2017grad} show that
SimCLR and DINOv3 may under-emphasize region and boundary cues, whereas
the adapted SAM encoder attends more strongly to the lesion structure.
\textbf{(b)} A Melanocytic Nevus (NV) and a Benign Keratosis-like lesion
(BKL) exhibit similar appearance
(SSIM~\cite{wang2004ssim} $=0.874$; visual-feature cosine similarity
$=0.952$), but distinct persistent homology signatures
(Wasserstein distance~\cite{cohen2005stability} $=0.417$).
The visual-only model misclusters the NV sample, whereas the
topology-guided visual model predicts it correctly.
Here, H$_0$ and H$_1$ summarize connected components and enclosed
structures, respectively.
}
\label{fig:motivation}

\end{figure}

Label-free clustering seeks to address the annotation bottleneck by
partitioning images directly from unlabeled data through joint
representation learning and grouping~\cite{ren2024deep,
zhou2024comprehensive,min2018survey}. In the dominant pseudo-labeling
paradigm~\cite{van2020scan,niu2022spice,jia2024towards}, a
self-supervised learning (SSL) backbone learns features from unlabeled
images~\cite{chen2020simple,chen2021mocov3}, and a clustering head
produces pseudo-labels that are iteratively refined
\cite{asano2020self}. However, existing methods still face two
limitations. 
\textit{(1) Limited use of structural priors:}
SSL-based methods, including SimCLR~\cite{chen2020simple},
MoCo-v3~\cite{chen2021mocov3}, BYOL~\cite{grill2020bootstrap}, and
DINOv3~\cite{simeoni2025dinov3}, learn strong visual representations
but may under-emphasize class-agnostic region and boundary cues that are
useful in medical imaging. As illustrated by the example in
Fig.~\ref{fig:motivation}(a), SAM~\cite{kirillov2023segment} provides
more explicit region- and boundary-sensitive features through its
mask-supervised pretraining.
\textit{(2) Absence of explicit topological encoding:}
Deep clustering methods such as SPICE~\cite{niu2022spice} and
CDC~\cite{jia2024towards}, as well as medical-specific methods such as
MedTrans~\cite{jahanian2024medtranscluster} and
DeepMCAT~\cite{kart2021deepmcat}, do not explicitly encode global
topological patterns such as connectivity, fragmentation, and enclosed
structures. Such patterns can provide useful information in medical
images~\cite{lawson2019persistent}. Figure~\ref{fig:motivation}(b)
illustrates one case in which visually similar samples exhibit
different persistent-homology signatures. These two representations are complementary but not equally discriminative. Visual-structural features retain rich appearance, semantic, and spatial information and therefore provide the primary clustering representation.
PH provides compact global geometric information for samples
that remain ambiguous in the visual feature space; we therefore use
topology to regularize rather than replace visual learning.

To address these limitations, we introduce VTRL-Med, a progressive
visual-structural learning framework with topology-guided
regularization for label-free medical image clustering. Specifically, Stage 1 adapts SAM to the medical domain while retaining structural cues
from its mask-supervised pretraining. Stage 2 produces reliable pseudo-labels via a two-step filtering process: Prototype-based selection retains high-confidence samples, and k-nearest neighbor (k-NN) based cross-view consistency filters locally inconsistent assignments. Stage 3 progressively
refines the visual representation after reliable pseudo-labels have
been established. During this refinement, an auxiliary topology branch
constructed from PH provides
complementary geometric regularization. The topology branch is used only during training, while final cluster assignments are produced by the refined visual branch alone. 

Our main contributions are three-fold. (1) We propose VTRL-Med, a progressive label-free medical image clustering framework that combines a primary visual-structural representation with complementary PH-based
regularization. (2) We develop a two-step pseudo-label filtering strategy and a progressive topology-guided refinement procedure that improves the visual representation. (3) We achieve SOTA performance on 13 medical image benchmarks, outperforming existing methods by 4.48\% in average accuracy. Topology-guided regularization further improves accuracy by 1.3--4.3\% over
visual refinement alone.

\section{Related Work}
\label{sec:related}

\noindent\textbf{Deep Clustering.}
Deep clustering jointly optimizes feature learning and cluster assignment~\cite{min2018survey,aljalbout2018clustering,yao2025learning,
yao2026reva,
yao2026semantic,zhu2024advancing,zhu2025fmri2ges,zhu2025pathology}.
Early methods (e.g., DEC~\cite{xie2016unsupervised}) alternate between representation learning and clustering with soft assignments.
Representative recent approaches adopt pseudo-labeling: SCAN~\cite{van2020scan} mines nearest neighbors for semantic consistency; SPICE~\cite{niu2022spice} uses prototype-based learning to sharpen inter-cluster separation; CDC~\cite{jia2024towards} calibrates confidence to improve pseudo-label quality.
For medical imaging, DeepMCAT~\cite{kart2021deepmcat} applies deep clustering to medical categorization with CNN features, and MedTrans~\cite{jahanian2024medtranscluster} leverages transfer learning followed by $k$-means. VTRL-Med differs by combining quota-based prototype selection with $k$-NN consistency filtering for reliable pseudo-labeling, followed by progressive visual refinement with auxiliary PH-based regularization.

\noindent\textbf{Self-Supervised Learning (SSL) for Medical Imaging.}
SSL learns representations from pretext tasks without labels.
Contrastive methods (SimCLR~\cite{chen2020simple}, MoCo-v3 \cite{chen2021mocov3}) maximize agreement between augmented views; BYOL~\cite{grill2020bootstrap} removes negatives with a momentum teacher; DINO/DINOv2/DINOv3~\cite{caron2021emerging,oquab2023dinov2,simeoni2025dinov3} employ self-distillation with ViTs (centering/sharpening) to obtain strong non-contrastive features.
While SSL has been adapted to medical imaging~\cite{Azizi_2021_ICCV}, domain-specific pretraining remains crucial.
SAM~\cite{kirillov2023segment}, trained on 11M images with 1.1B masks, learns boundary-aware, class-agnostic segmentation cues; SAM2~\cite{SAM2} extends this framework to videos; MedSAM~\cite{MedSAM} and MedSAM2~\cite{MedSAM2} tailor the paradigm to medical segmentation tasks.

\noindent\textbf{Persistent Homology in Medical Imaging.}
Appearance-driven DL methods can overlook global geometric properties relevant to some medical imaging tasks~\cite{peng2024phg,gu2026integrating,adame2025topo,gu2025self,meng2026topoagent} (e.g., connectivity of tissue components, glandular lumens, and cell distributions~\cite{elston1991pathological,lawson2019persistent}).
Persistent homology (PH)~\cite{edelsbrunner2010computational} quantifies topology across scales by tracking births and deaths of connected components ($H_0$) and loops ($H_1$), summarizing them in persistence diagrams (PDs) with stability guarantees~\cite{cohen2005stability,chambers2025stable}.
We employ \emph{cubical} PH~\cite{wagner2011efficient}, which builds filtrations directly from pixel features and efficiently extracts topological signatures without explicit segmentation.
Prior work applied PH to tumor analysis~\cite{qaiser2019fast}, cancer grading~\cite{lawson2019persistent,singh2014topological,gu2025topoimages}, and supervised classification (e.g., PHG-Net~\cite{peng2024phg}), predominantly in \emph{labeled} settings.

\section{Method}
\label{sec:method}

\begin{figure*}[t]
    \centering
\includegraphics[width=0.85\textwidth,trim={1.6cm 6.2cm 3cm 2.5cm},clip]{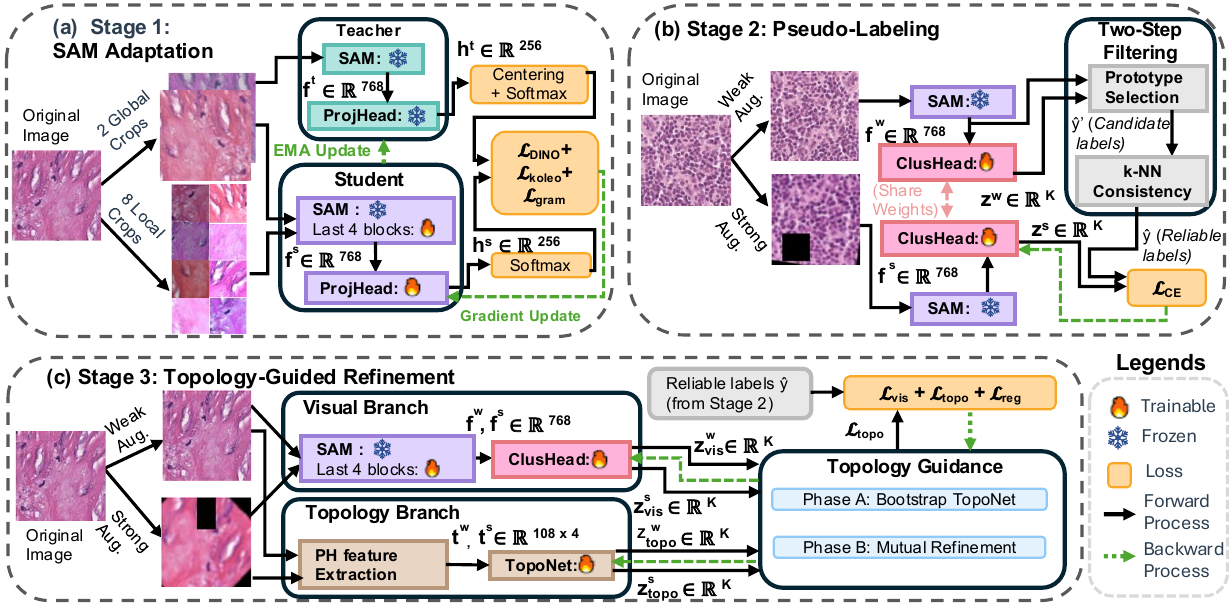}

\caption{Illustrating the three-stage VTRL-Med framework.
\textbf{(a) Stage 1:} Adapting the SAM image encoder to extract
domain-specific visual features and structure-sensitive cues inherited
from mask-supervised pretraining.
\textbf{(b) Stage 2:} Training the clustering head (ClusHead) with
confidence-and-consistency filtering to produce reliable pseudo-labels.
\textbf{(c) Stage 3:} Progressively refining the visual branch with
auxiliary topology guidance through bootstrap (Phase A) and mutual
refinement (Phase B). $K$ denotes the number of clusters.}

\label{fig:framework}

\end{figure*}

Our \textbf{VTRL-Med} is a three-stage framework for \rev{label-free} medical image clustering. Given an unlabeled dataset of medical images, $\mathcal{D}=\{\mathbf{x}_i\}_{i=1}^N$, the goal is to partition its samples into $K$ clusters. Our design is motivated by the observation that effective medical image clustering requires three complementary mechanisms: (i) structural priors that anchor features to anatomically meaningful regions, (ii) reliable pseudo-labels that bootstrap cluster assignments, and (iii) topological priors that resolve ambiguities at cluster boundaries where visual features alone are insufficient. Each stage of VTRL-Med focuses on one of these mechanisms (see Fig.~\ref{fig:framework}). \textbf{Stage 1} adapts a SAM image encoder to the medical domain to produce domain-specific visual features and mask-based structural priors (Section~\ref{sec:stage1}); \textbf{Stage 2} trains a clustering head and generates reliable pseudo-labels via a two-step filtering (Section~\ref{sec:stage2}); \textbf{Stage 3} progressively refines the visual representation after these reliable pseudo-labels have been established. During this refinement, an auxiliary topology branch provides training-time geometric regularization
(Section~\ref{sec:stage3}). Final cluster assignments are produced by
the refined visual branch alone.

\subsection{Stage 1: SAM Adaptation for Visual Features \& Mask-Based Structural Priors}
\label{sec:stage1}

The performance of deep clustering critically depends on discriminative feature representations that capture both semantic content and structural characteristics. To attain a robust visual feature backbone that captures structural priors, Stage 1 adapts SAM~\cite{kirillov2023segment} to the medical image domain. We leverage SAM because its zero-shot performance on segmentation tasks provides class-agnostic structural priors (boundary/region cues) that are useful for medical image analysis (Fig.~\ref{fig:motivation}(a)). However, since SAM is pretrained on natural images, the domain gap results in suboptimal features for medical images. To address this challenge, we employ a DINOv3-inspired self-supervised adaptation~\cite{simeoni2025dinov3} that balances domain adaptation and structural prior preservation.

We employ a teacher-student model (Fig.~\ref{fig:framework}(a)) in which both networks share the SAM image encoder architecture. The student network fine-tunes only the last $L=4$ Transformer blocks while freezing all earlier layers to balance adaptation and computational efficiency. A projection head (a 3-layer MLP: $768 \rightarrow 2048 \rightarrow 2048 \rightarrow 256$) maps the globally pooled SAM features $\mathbf{f} \in \mathbb{R}^{768}$ to normalized embeddings $\mathbf{h} \in \mathbb{R}^{256}$ for self-distillation. The teacher network receives no gradient updates and is updated via exponential moving average (EMA) of the student parameters \cite{caron2021emerging}, providing stable training targets throughout adaptation. 
The adaptation optimizes three objectives:
$\mathcal{L}_{\text{Stage1}} = \mathcal{L}_{\text{DINO}} +  \mathcal{L}_{\text{koleo}} +  \mathcal{L}_{\text{gram}},$
where $\mathcal{L}_{\text{DINO}}$ enforces consistency between the student and teacher predictions via centering and temperature-scaled cross-entropy~\cite{caron2021emerging}, $\mathcal{L}_{\text{koleo}}$ is an entropy regularizer that promotes uniform feature distribution to prevent representation collapse~\cite{simeoni2025dinov3}, and $\mathcal{L}_{\text{gram}}$ preserves spatial structures through Gram matrix anchoring~\cite{simeoni2025dinov3}. Full schedules and hyperparameters are
provided in the supplementary material.

This three-step process yields $\mathcal{F}^*$, a medical-domain-adapted SAM image encoder that serves as the feature backbone for downstream clustering tasks.

\subsection{Stage 2: Confidence-and-Consistency Pseudo-Labeling}
\label{sec:stage2}

Stage 2 generates reliable pseudo-labels for bootstrapping the clustering head. We adopt a weak-strong augmentation strategy~\cite{sohn2020fixmatch}, which processes each image with two different augmentation strengths (see Fig.~\ref{fig:framework}(b)). We freeze the encoder $\mathcal{F}^*$ to extract features from both augmented views while preventing catastrophic forgetting of Stage 1's learned structural priors. A trainable clustering head (ClusHead $\mathcal{C}$: a 2-layer MLP, 768$\rightarrow$512$\rightarrow$$K$) then maps these features to cluster logits $\mathbf{z}_i^w, \mathbf{z}_i^s \in \mathbb{R}^K$, where $K$ is the number of clusters. To ensure pseudo-label quality, we apply a two-step filtering process on the weak-augmentation predictions to identify reliable labels, which subsequently supervise ClusHead training on the strong augmentations.

A key challenge in Stage 2 is to balance coverage and the quality of the pseudo-labels: Strict confidence thresholds yield few pseudo-labels and limited training signals, while relaxed thresholds introduce noise that degrades clustering. To address this trade-off, we propose a two-step filtering that combines prototype-based selection with k-NN consistency validation. Our idea is that prototype-based selection with quota ensures sufficient coverage by guaranteeing samples from each cluster, while k-NN validation improves pseudo-label quality by removing geometrically inconsistent outliers, together achieving better coverage-quality balance than single-threshold schemes. \rev{The specific thresholds (top 50\%, $M = \lfloor N/(2K) \rfloor$, 70\% k-NN agreement) are selected on three tuning datasets; sensitivity analysis (Section~\ref{sec:ablation}) confirms robustness within $\pm$1.5\% accuracy.}

\begin{algorithm}[t]
\scriptsize
\caption{Two-Step Filtering}
\label{alg:dual_step_filtering}
\begin{algorithmic}[1]
\REQUIRE Features $\{\mathbf{f}_i^w\}_{i=1}^N$, $\mathbf{f}_i^w \!\in\! \mathbb{R}^{768}$; logits $\{\mathbf{z}_i^w\}_{i=1}^N$, $\mathbf{z}_i^w \!\in\! \mathbb{R}^K$
\ENSURE Pseudo-labels $\{\hat{y}_i\}_{i=1}^N$, $\hat{y}_i \in \{1,\ldots,K,-1\}$
\STATE Compute $p_{ik} = \text{softmax}(\mathbf{z}_i^w)_k$ $\forall\, i,k$;\; set $M = \lfloor N/(2K)\rfloor$
\STATE Init $\hat{y}'_i = -1$, $s^*_i = -\infty$ $\forall\, i$
\STATE \textbf{Step 1: Prototype Selection}
\FOR{$k = 1, \ldots, K$}
    \STATE $\mathcal{S}_k \!\leftarrow\! \{i : \arg\max_j p_{ij} = k\}$;\; select top 50\% of $\mathcal{S}_k$ by $p_{ik} \!\rightarrow\! \mathcal{H}_k$
    \STATE $\boldsymbol{\mu}_k = \frac{1}{|\mathcal{H}_k|}\sum_{i \in \mathcal{H}_k} \mathbf{f}_i^w$
\ENDFOR
\STATE Compute $s_{ik} = \text{cosine}(\mathbf{f}_i^w, \boldsymbol{\mu}_k)$ $\forall\, i, k$
\FOR{$k = 1, \ldots, K$}
    \STATE Select top $M$ by $s_{ik} \!\rightarrow\! \mathcal{A}_k$;\; \textbf{for} $i \!\in\! \mathcal{A}_k$ \textbf{do if} $s_{ik} > s^*_i$: $s^*_i \!\leftarrow\! s_{ik}$, $\hat{y}'_i \!\leftarrow\! k$
\ENDFOR
\STATE \textbf{Step 2: k-NN Consistency}
\FOR{$i$ with $\hat{y}'_i \neq -1$}
    \STATE Find 10 nearest neighbors $\mathcal{N}_i$ of $\mathbf{f}_i^w$ by cosine distance
    \STATE $\alpha_i = \frac{1}{10}|\{j \!\in\! \mathcal{N}_i : \hat{y}'_j = \hat{y}'_i\}|$;\; $\hat{y}_i \!\leftarrow\! \hat{y}'_i$ if $\alpha_i \!\geq\! 0.7$, else $-1$
\ENDFOR
\end{algorithmic}

\end{algorithm}

Specifically, as shown in Algorithm~\ref{alg:dual_step_filtering}, Step 1 first computes cluster prototypes: For each cluster $k$, we identify high-confidence samples (top 50\% by prediction confidence) and compute the prototype $\mathbf{\mu}_k$ as their mean features (lines 5--6). We then compute cosine similarity between all samples and prototypes (line 8), and assign candidate labels using a quota-based strategy: Each cluster receives up to $M = \lfloor N/(2K) \rfloor$ samples with the highest similarity to its prototype (lines 9--11). If a sample appears in multiple clusters' top-$M$ sets, it is assigned to the cluster with the maximum similarity, ensuring each sample receives its most confident assignment. Step 2 validates the candidates by requiring 70\% label agreement among their 10 nearest neighbors in the feature space (lines 13--16), producing final reliable labels $\hat{y}$, where $-1$ marks filtered samples.

In each training iteration, weak features $\{\mathbf{f}_i^w\}_{i=1}^N$ pass through ClusHead to attain logits $\{\mathbf{z}_i^w\}_{i=1}^N$, which are filtered via Algorithm~\ref{alg:dual_step_filtering} to produce reliable labels. The labels then supervise the strong-view logits $\{\mathbf{z}_i^s\}_{i=1}^N$ (obtained by passing $\{\mathbf{f}_i^s\}_{i=1}^N$ in ClusHead) with cross-entropy loss:
$\mathcal{L}_{\text{Stage2}} = \frac{1}{|\mathcal{R}|}\sum_{i \in \mathcal{R}} \ell_{\text{CE}}(\mathbf{z}_i^s, \hat{y}_i),$
where $\mathcal{R} = \{i \mid \hat{y}_i \neq -1\}$ is the reliable set and $\ell_{\text{CE}}$ is cross-entropy loss. Stage 2 outputs the trained ClusHead $\mathcal{C}$, reliable pseudo-labels $\{\hat{y}_i\}$, and reliable set $\mathcal{R}$.

\subsection{Stage 3: Progressive Visual Refinement with Topological Regularization}
\label{sec:stage3}

Stage 2 establishes initial cluster assignments using reliable pseudo-labels. Nevertheless, residual errors may remain when samples have similar visual appearance but different global geometric
organization, as illustrated in Fig.~\ref{fig:motivation}(b).
Stage 3 therefore progressively refines the primary visual branch while
an auxiliary topology branch provides complementary training-time
regularization. The topology branch is intended to supplement, rather
than replace, the more discriminative visual representation.

Specifically, both weak and strong augmented views pass through the two-branch network in parallel (see Fig.~\ref{fig:framework}(c)). The visual branch consists of the adapted SAM encoder $\mathcal{F}^*$ (the last 4 blocks are fine-tunable) and ClusHead $\mathcal{C}$ from Stages 1-2, producing logits $\mathbf{z}_{\text{vis}}^w, \mathbf{z}_{\text{vis}}^s \in \mathbb{R}^K$. The topology branch processes PH features from both augmented images: We compute cubical complex filtration to extract the top 36 H$_0$ (connected components) and top 72 H$_1$ (topological holes) based on persistence thresholding to retain salient structures while filtering noise, yielding $\mathbf{t}^w, \mathbf{t}^s \in \mathbb{R}^{108 \times 4}$; each feature is represented by (birth time, death time, persistence, homology dimension). These topological representations capture geometric structures useful for medical imaging: H$_0$ quantifies tissue fragmentation patterns and H$_1$ detects enclosed structures (e.g., glandular lumens~\cite{lawson2019persistent,qaiser2019fast}). \rev{While standard PH discards the spatial locations of 
topological features, for image-level clustering the 
presence and multiplicity of structures matter more 
than their pixel coordinates.} \rev{PH features and SAM features are complementary by 
design: SAM encodes appearance features and spatial 
structures, while PH summarizes geometric structures 
(component counts, loop structures) that are invariant 
to continuous deformations of the image domain, capturing 
global topological distinctions that local appearance 
features may miss.} TopoNet processes topological representations $\mathbf{t}^w, \mathbf{t}^s$ using three 1D convolutional layers (kernel size 3) with 128, 256, and 512 filters, each followed by ReLU activation. Global max-pooling aggregates spatial features into 512-D vectors, which are mapped to cluster predictions via a three-layer MLP (512$\rightarrow$256$\rightarrow$128$\rightarrow$K) to produce topological logits $\mathbf{z}_{\text{topo}}^w, \mathbf{z}_{\text{topo}}^s \in \mathbb{R}^K$.

Given the outputs from both branches and reliable pseudo-labels $\{\hat{y}_i\}$ from Stage 2, we optimize a unified training objective:
\begin{equation}
\mathcal{L}_{\text{Stage3}} = \mathcal{L}_{\text{vis}} + \lambda_{\text{topo}} \mathcal{L}_{\text{topo}} + \lambda_{\text{reg}} \mathcal{L}_{\text{reg}},
\label{eq:stage3_loss}
\end{equation}
where $\lambda_{\text{topo}}$ is the phase-dependent topology branch weight, and $\lambda_{\text{reg}} = 0.1$ is the regularization weight. The visual loss $\mathcal{L}_{\text{vis}}$ trains the visual branch using reliable pseudo-labels and high-confidence predictions from both augmented views:
\begin{equation}
\mathcal{L}_{\text{vis}} = \frac{1}{|\mathcal{R}|}\sum_{i \in \mathcal{R}} \ell_{\text{CE}}(\mathbf{z}_{\text{vis},i}^w, \hat{y}_i) + \frac{1}{N}\sum_{i=1}^N m_i \cdot \ell_{\text{CE}}(\mathbf{z}_{\text{vis},i}^s, \tilde{y}_i),
\label{eq:lvis}
\end{equation}
where $\tilde{y}_i = \arg\max_j (\mathbf{z}_{\text{vis},i}^w)_j$ are weak-view pseudo-labels, $m_i = \mathbbm{1}\{\max_j (\mathbf{p}_{\text{vis},i}^w)_j \geq \tau\}$ selects high-confidence predictions with $\mathbf{p}_{\text{vis},i}^w = \text{softmax}(\mathbf{z}_{\text{vis},i}^w)$ for each sample $i$, $\tau = 0.8$, and $\mathcal{R}$ is the reliable set from Stage 2. Following~\cite{caron2020unsupervised}, $\mathcal{L}_{\text{reg}}$ prevents cluster collapse through cluster balance, entropy maximization, and swapped prediction losses.

While $\mathcal{L}_{\text{vis}}$ remains fixed across \rev{both} training phases, a critical challenge is to design $\mathcal{L}_{\text{topo}}$ for training the topology branch and coordinating interactions between the two branches. This challenge arises from the mismatched initialization quality: The visual branch inherits Stage 1's adapted SAM encoder and Stage 2's trained ClusHead, providing strong initial clustering performance. In contrast, the topology branch with a random-initialized TopoNet starts from scratch. Naively enforcing agreement between these two would propagate topology errors to the well-performing visual branch, degrading the overall clustering quality. To alleviate this initialization mismatch, we propose a progressive topology guidance strategy with phase-dependent \rev{loss $\mathcal{L}_{\text{topo}}$ and weight $\lambda_{\text{topo}}$ that control their interactions: Phase A bootstraps TopoNet independently; Phase B enables mutual refinement once both branches are sufficiently mature}.

In Phase A (Bootstrap TopoNet), both branches learn independently from the reliable pseudo-labels to establish TopoNet's basic clustering capability:
$\mathcal{L}_{\text{topo}}^A = \frac{1}{|\mathcal{R}|}\sum_{i \in \mathcal{R}} \ell_{\text{CE}}(\mathbf{z}_{\text{topo},i}^w, \hat{y}_i) + \frac{1}{N}\sum_{i=1}^N m_i \cdot \ell_{\text{CE}}(\mathbf{z}_{\text{topo},i}^s, \tilde{y}_i),$
mirroring $\mathcal{L}_{\text{vis}}$ in Eq.~(\ref{eq:lvis}), weighted by $\lambda_{\text{topo}} ^A = 0.5$ in Eq.~(\ref{eq:stage3_loss}).

Once TopoNet stabilizes, Phase B (Mutual Refinement) enables bidirectional learning where both branches refine each other with high-confidence predictions: $\mathcal{L}_{\text{topo}}^B = \mathcal{L}_{\text{consist}} + \mathcal{L}_{\text{mutual}},$ where $\mathcal{L}_{\text{consist}} = \tfrac{1}{|\mathcal{Q}|}\textstyle\sum_{i \in \mathcal{Q}} \ell_{\text{CE}}\!\left(\mathbf{z}_{\text{topo},i}^s,\, \arg\max_j (\mathbf{z}_{\text{vis},i}^s)_j\right)$ and $\mathcal{L}_{\text{mutual}} = \tfrac{1}{|\mathcal{A}|}\textstyle\sum_{i \in \mathcal{A}} D_{\text{KL}}\!\left(\mathbf{p}_{\text{vis},i}^s \| \mathbf{p}_{\text{topo},i}^s\right).$
%
Here, $\mathcal{L}_{\text{consist}}$ encourages TopoNet to learn from confident visual predictions with $\mathcal{Q} = \{i \mid \max_j (\mathbf{p}_{\text{vis},i}^s)_j \geq \tau\}$. $\mathcal{L}_{\text{mutual}}$ aligns the two branches on samples when both agree with high confidence, where $\mathbf{p}_{\text{vis},i}^s = \text{softmax}(\mathbf{z}_{\text{vis},i}^s)$ and $\mathbf{p}_{\text{topo},i}^s = \text{softmax}(\mathbf{z}_{\text{topo},i}^s)$ denote the predicted probabilities from the visual branch and TopoNet for each strong augmented sample $i$, respectively. $D_{\text{KL}}$ denotes KL divergence, and $\mathcal{A}$ includes samples on which both predict the same class with confidence $\geq \tau$. $\lambda_{\text{topo}} ^B = 0.2$ in Eq.~(\ref{eq:stage3_loss}) for Phase B. During inference, only the visual branch  is used for cluster assignments; the topology branch serves solely as training-time regularization, adding no inference cost.


\begin{table}[bt]
\centering
\caption{Dataset statistics of the 13 medical image benchmarks. 
MedMNIST v2 images~\cite{medmnistv2} are natively $224 \times 224$; ISIC2019 and Kvasir images are resized to $224 \times 224$.}
\label{tab:datasets}
\scriptsize

\begin{tabular}{llcc}
\toprule
\textbf{Dataset} & \textbf{Modality} & \textbf{Classes} & \textbf{Train / Test} \\
\midrule
PathMNIST & Colon Pathology & 9 & 89,996 / 7,180 \\
DermaMNIST & Dermatoscopy & 7 & 7,007 / 2,005 \\
OCTMNIST & Retinal OCT & 4 & 97,477 / 1,000 \\
PneumoniaMNIST & Chest X-ray & 2 & 4,708 / 624 \\
RetinaMNIST & Fundus Photography & 5 & 1,080 / 400 \\
BreastMNIST & Breast Ultrasound & 2 & 546 / 156 \\
BloodMNIST & Blood Cell Microscopy & 8 & 11,959 / 3,421 \\
TissueMNIST & Kidney Histology & 8 & 165,466 / 47,280 \\
OrganAMNIST & Abdominal CT (Axial) & 11 & 34,581 / 17,778 \\
OrganCMNIST & Abdominal CT (Coronal) & 11 & 13,000 / 8,268 \\
OrganSMNIST & Abdominal CT (Sagittal) & 11 & 13,940 / 8,829 \\
\midrule
ISIC2019 & Skin Lesion Dermoscopy & 8 & 25,331 / 8,238 \\
Kvasir & GI Endoscopy & 8 & 3,200 / 800 \\
\bottomrule
\end{tabular}

\end{table}

\section{Experiments}

\textbf{Datasets.} We evaluate on 13 medical image datasets, spanning diverse anatomical regions, imaging modalities, and scales (see Table~\ref{tab:datasets}): 11 datasets from MedMNIST v2~\cite{medmnistv1,medmnistv2}, plus ISIC2019~\cite{tschandl2018ham10000} and Kvasir~\cite{pogorelov2017kvasir}.

\noindent\textbf{Methods.} We compare VTRL-Med (trained for 300 epochs: Stage 1, 150 epochs; Stage 2, 50 epochs; Stage 3, 100 epochs) with: (1) SSL-based clustering (SimCLR~\cite{chen2020simple}, MoCo-v3~\cite{chen2021mocov3}, BYOL~\cite{grill2020bootstrap}, DINOv3~\cite{simeoni2025dinov3}) trained for 200 epochs (sufficient for convergence on the medical datasets) followed by $K$-means; (2) deep clustering (SPICE~\cite{niu2022spice}, CDC~\cite{jia2024towards}); (3) medical-specific methods (MedTrans~\cite{jahanian2024medtranscluster}, DeepMCAT \cite{kart2021deepmcat}); (4) a SOTA supervised method MedMamba-B~\cite{yue2024medmamba} (as an upper bound). \rev{All methods rely on pretrained backbones (ImageNet~\cite{deng2009imagenet} for SSL methods, SA-1B~\cite{kirillov2023segment} for SAM); no method uses target-task annotations.} Supplementary 6 gives all training details.

\noindent
\rev{\textbf{Tuning Protocol.} All hyperparameters, for both VTRL-Med and all baselines, were selected to maximize average performance across three tuning datasets (BloodMNIST, TissueMNIST, OrganAMNIST) and applied unchanged to the remaining 10 benchmarks. For the baselines, we re-implemented each one using official codebases and tuned the learning rate, batch size, and method-specific parameters on the same tuning sets. No per-dataset tuning was performed for any method. Sensitivity analysis (Section~\ref{sec:ablation}) confirms that the performance is stable within $\pm$1.5\% across all tested hyperparameter ranges.}

\noindent\textbf{Evaluation Metrics.} Following~\cite{niu2022spice,jia2024towards,meng2025efficient,meng2026psychology}, all methods use $K$ as the cluster number.
We report ACC (Hungarian matching~\cite{kuhn1955hungarian}), NMI~\cite{strehl2002cluster}, and ARI~\cite{hubert1985comparing}, averaged over 3 runs. Statistical significance is assessed via paired t-tests. All experiments are conducted on an NVIDIA H100 GPU.

\begin{table*}[ht!]
\centering
\caption{Clustering performances of different methods on 13 medical image datasets. The best and second-best \rev{label-free} results are \textbf{bold} and \underline{underlined}. \rev{MedMamba is a SOTA supervised method served as an upper bound; all other methods are label-free.} Standard deviations and statistical tests are in Supplementary 7 and 8.}
\label{tab:main_results}
\scriptsize
\setlength{\tabcolsep}{1pt}
\resizebox{\textwidth}{!}{%
\begin{tabular}{lccccccccccccccccccccc}
\toprule
& \multicolumn{3}{c}{\textbf{PathMNIST}} & \multicolumn{3}{c}{\textbf{DermaMNIST}} & \multicolumn{3}{c}{\textbf{OCTMNIST}} & \multicolumn{3}{c}{\textbf{PneumoniaMNIST}} & \multicolumn{3}{c}{\textbf{RetinaMNIST}} & \multicolumn{3}{c}{\textbf{BreastMNIST}} & \multicolumn{3}{c}{\textbf{BloodMNIST}} \\
\cmidrule(lr){2-4}\cmidrule(lr){5-7}\cmidrule(lr){8-10}\cmidrule(lr){11-13}\cmidrule(lr){14-16}\cmidrule(lr){17-19}\cmidrule(lr){20-22}
\textbf{Method} & ACC & NMI & ARI & ACC & NMI & ARI & ACC & NMI & ARI & ACC & NMI & ARI & ACC & NMI & ARI & ACC & NMI & ARI & ACC & NMI & ARI \\
\midrule
SimCLR & 77.21 & 83.51 & 75.66 & 26.48 & 16.01 & 5.73 & 76.10 & 53.23 & 50.94 & 73.88 & 20.42 & 22.68 & 31.50 & 8.30 & 5.03 & 64.74 & 10.37 & 8.20 & 69.22 & 70.70 & 60.99\\
MoCo-v3 & 77.53 & 82.36 & 74.08 & 36.06 & 14.29 & 9.72 & 87.00 & 68.71 & 70.03 & 75.64 & 26.04 & 26.14 & 27.75 & 5.36 & 0.56 & 57.05 & 1.73 & 4.55 & 89.39 & 84.84 & 82.53 \\
BYOL & 77.56 & 83.13 & 74.80 & 56.94 & 13.07 & 16.05 & 88.50 & 71.18 & 72.85 & 86.70 & 43.32 & 53.74 & 35.75 & 13.76 & 5.12 & 54.49 & 2.99 & 1.28 & 86.73 & 81.58 & 79.69 \\
DINOv3 &76.42 &81.23 &77.24 &59.63 &30.54 &34.09 &85.50 &\textbf{80.86} &\textbf{80.83} &\textbf{91.99} &\underline{58.84} &\underline{70.29} &42.50 &15.45 &10.53 &60.00 &\underline{16.23} &14.67 &90.48 &\underline{88.18} &80.83 \\
\midrule
SPICE & \underline{93.72} & \underline{88.35} & \underline{90.13} & \underline{70.62} & \underline{36.09} & \underline{44.29} & 78.20 & 64.48 & 56.85 & 88.75 & \textbf{62.18} & \textbf{72.38} & 43.75 & 9.39 & 16.08 & \underline{76.92} & 11.51 & \underline{23.04} & 88.52 & 82.19 & 83.14 \\
CDC & 81.99 & 83.95 & 83.29 & 53.77 & 33.51 & 29.10 & \underline{90.60} & \underline{76.49} & \underline{77.62} & 85.16 & 46.39 & 45.13 & 44.50 & 16.68 & 14.32 & 61.92 & 13.94 & 11.94 & \underline{91.55} & 87.18 & \underline{84.92} \\
MedTrans & 74.86 & 74.66 & 57.42 & 53.64 & 24.12 & 23.58 & 72.33 & 56.76 & 53.67 & 79.85 & 41.34 & 35.59 & \textbf{52.50} & \underline{28.12} & \underline{28.17} & 70.51 & 13.14 & 16.34 & 77.39 & 67.86 & 65.84 \\
DeepMCAT & 76.92 & 67.19 & 64.31 & 59.57 & 25.39 & 23.41 & 83.17 & 62.21 & 51.89 & 85.54 & 51.34 & 51.25 & 43.25 & \textbf{29.99} & 26.21 & 72.44 & 10.64 & 9.08 & 85.95 & 79.11 & 76.28 \\
\midrule
VTRL-Med &\textbf{96.22} &\textbf{92.64} &\textbf{93.77} & \textbf{74.29} &\textbf{49.59} &\textbf{59.71} &\textbf{91.32} &74.16 &73.61 &\underline{90.71} &52.04 &49.79 &\underline{51.00} &27.39 &\textbf{33.40} & \textbf{81.82} & \textbf{34.59} &\textbf{50.08} &\textbf{98.83} &\textbf{96.81} &\textbf{97.05} \\
\midrule
MedMamba & 96.93 & 93.51 & 94.61 & 75.60 & 62.80 & 60.50 & 92.20 & 79.02 & 73.71 & 94.22 & 51.29 & 57.70 & 54.25 & 18.98 & 25.04 & 85.44 & 23.12 & 30.97 & 99.30 & 95.36 & 95.96 \\
\bottomrule
\end{tabular}%
}

\resizebox{\textwidth}{!}{%
\begin{tabular}{lccccccccccccccccccccc}
\toprule
& \multicolumn{3}{c}{\textbf{TissueMNIST}} & \multicolumn{3}{c}{\textbf{OrganAMNIST}} & \multicolumn{3}{c}{\textbf{OrganCMNIST}} & \multicolumn{3}{c}{\textbf{OrganSMNIST}} & \multicolumn{3}{c}{\textbf{ISIC2019}} & \multicolumn{3}{c}{\textbf{Kvasir}} & \multicolumn{3}{c}{\textbf{Average}} \\
\cmidrule(lr){2-4}\cmidrule(lr){5-7}\cmidrule(lr){8-10}\cmidrule(lr){11-13}\cmidrule(lr){14-16}\cmidrule(lr){17-19}\cmidrule(lr){20-22}
\textbf{Method} & ACC & NMI & ARI & ACC & NMI & ARI & ACC & NMI & ARI & ACC & NMI & ARI & ACC & NMI & ARI & ACC & NMI & ARI & ACC & NMI & ARI \\
\midrule
SimCLR & 29.09 & 21.38 & 11.60 & 59.64 & 66.25 & 47.82 & 58.26 & 61.82 & 43.80 & 47.06 & 50.09 & 31.19 & 23.61 & 11.82 & 6.17 & 54.33 & 54.63 & 40.24 & 53.16 & 40.66 & 31.54 \\
MoCo-v3 & 27.73 & 19.08 & 9.20 & 48.45 & 56.91 & 34.51 & 39.24 & 45.18 & 23.62 & 34.82 & 39.06 & 17.71 & 27.49 & 9.27 & 6.99 & \textbf{79.67} & \textbf{71.76} & \textbf{63.30} & 54.45 & 40.35 & 32.53 \\
BYOL& 29.99 & 20.06 & 11.97 & 51.04 & 57.67 & 33.64 & 39.34 & 46.18 & 25.81 & 34.12 & 41.69 & 22.26 & 22.71 & 12.59 & 6.98 & 60.33 & 63.23 & 48.14 & 55.71 & 42.34 & 34.79 \\
DINOv3 &28.68 &19.42 &9.29 &60.96 &63.48 &68.67 &53.14 &53.36 &40.96 &36.67 &44.54 &27.48 &39.82 &14.67 &19.84 &65.62 &63.37 &53.04 & 60.88 & 48.47 & 45.21 \\
\midrule
SPICE & \underline{70.22} & \textbf{42.44} & \textbf{47.96} & \underline{88.74} & \underline{80.86} & \underline{77.39} & \underline{84.82} & 77.23 & \underline{77.36} & \underline{72.00} & 63.58 & 57.17 & \underline{67.00} & \underline{31.19} & \textbf{45.95} & \underline{78.17} & \underline{69.11} & \underline{60.96} &  \underline{77.03} &  \underline{55.28}  &  \underline{57.90} \\
CDC & 61.11 & 30.35 & 31.02 & 80.86 & 79.31 & 67.86 & 76.81 & 76.85 & 69.13 & 70.85 & 67.68 & \underline{58.74} & 51.47 & 26.85 & 30.62 & 63.69 & 54.75 & 48.18 & 70.33 & 53.38 & 50.14 \\
MedTrans & 52.29 & 22.12 & 24.69 & 78.08 & 79.89 & 65.42 & 73.49 & \underline{79.65} & 71.40 & 63.83 & \textbf{71.16} & 57.17 & 40.99 & 19.06 & 18.46 & 56.50 & 60.77 & 45.33 & 65.10 & 49.13 & 43.31 \\
DeepMCAT & 57.68 & 28.42 & 35.61 & 82.37 & 68.52 & 44.19 & 66.56 & 60.19 & 59.37 & 60.27 & 58.91 & 53.83 & 52.15 & 14.38 & 16.52 & 57.17 & 50.62 & 40.83 & 67.93 & 46.69 & 42.52 \\
\midrule
VTRL-Med &\textbf{70.37} &\underline{41.83} &\underline{47.65} & \textbf{90.73} & \textbf{84.99} & \textbf{83.53} & \textbf{90.17} &\textbf{82.18} & \textbf{83.19} & \textbf{75.50} & \underline{68.67} & \textbf{68.83} & \textbf{71.51} & \textbf{47.90} & \underline{39.43} & 77.17 & 63.31 & 43.35 & \textbf{81.51} & \textbf{62.78} & \textbf{63.34} \\
\midrule
MedMamba & 71.34 & 43.64 & 50.33 & 95.93 & 88.18 & 88.62 & 94.45 & 86.94 & 89.57 & 83.81 & 73.26 & 74.50 & 76.15 & 54.39 & 55.67 & 82.83 & 75.67 & 67.12 & 84.80  & 65.09 & 66.48 \\
\bottomrule
\end{tabular}%
}

\end{table*}

\subsection{Main Results}
\label{sec:results}
Table~\ref{tab:main_results} reports clustering performance on all 13 datasets. VTRL-Med achieves SOTA performance among \rev{label-free} methods, ranking first on 10/13 datasets with average ACC/NMI/ARI of 81.51\%/62.78\%/63.34\%. Compared to the 2nd-best baseline SPICE, VTRL-Med improves ACC by 4.48\%, NMI by 7.50\%, and ARI by 5.44\%, with lower standard deviation (1.13\% vs 1.34\%). All improvements are statistically significant ($p{<}0.02$; see Supplementary 8).
VTRL-Med closely approaches the supervised performance, with only a 3.29\% ACC gap to MedMamba-B ($p{=}0.021$). On four datasets (PathMNIST, BloodMNIST, TissueMNIST, OCTMNIST), \rev{VTRL-Med} attains near-supervised performance with $<$1\% ACC gaps ($p{<}0.05$), indicating that \rev{label-free} clustering can match supervised methods when visual patterns are highly discriminative.
VTRL-Med underperforms on three datasets: RetinaMNIST (where MedTrans leads by 1.50\% in ACC), Kvasir (where MoCo-v3 leads by 2.50\%), and PneumoniaMNIST (where DINOv3 leads by 1.28\%). These gaps are comparable to standard deviations, suggesting similar performance within the statistical uncertainty.

\subsection{Component Analysis and Ablation}
\label{sec:ablation}

\begin{table}[bt]
\centering

\caption{Component analysis and ablation on 3 
datasets. Rows 1--3: stage necessity; rows 4--7: 
Stage 1 design; rows 8--15: Stage 2 design; rows 
16--21: Stage 3 design and full model. $^\dagger$SAM adapted features 
with pseudo-labeling only (no Stage 3 refinement).}
\label{tab:ablation_study}
\scalebox{0.62}{
\begin{tabular}{@{}l c@{\,/\,}c@{\,/\,}c@{\quad} c@{\,/\,}c@{\,/\,}c@{\quad} c@{\,/\,}c@{\,/\,}c@{}}
\toprule
& \multicolumn{3}{c}{\textbf{PathMNIST}} & \multicolumn{3}{c}{\textbf{ISIC2019}} & \multicolumn{3}{c}{\textbf{Kvasir}} \\
\textbf{Configuration} & \multicolumn{3}{c}{\textbf{ACC/NMI/ARI}} & \multicolumn{3}{c}{\textbf{ACC/NMI/ARI}} & \multicolumn{3}{c}{\textbf{ACC/NMI/ARI}} \\
\midrule
w/o Stage 1            & 56.91 & 36.18 & 40.23 & 36.76 & 17.59 & 20.53 & 47.37 & 43.04 & 28.91 \\
w/o Stage 2            & 92.58 & 88.52 & 89.51 & 55.03 & 22.37 & 35.04 & 69.29 & 50.31 & 48.18 \\
w/o Stage 3            & 63.69 & 58.02 & 47.88 & 51.71 & 20.40 & 19.23 & 68.15 & 44.41 & 31.09 \\
\midrule
Raw pixels             & 45.11 & 45.37 & 30.62 & 23.35 &  6.93 &  5.29 & 45.78 & 45.35 & 29.22 \\
ResNet-50 (ImageNet)   & 73.82 & 74.29 & 67.24 & 22.66 &  5.41 &  3.67 & 55.28 & 59.56 & 43.89 \\
SAM ViT-B (frozen)     & 47.90 & 50.13 & 31.17 & 25.13 &  6.32 &  7.04 & 49.51 & 46.19 & 31.32 \\
SAM ViT-B (adapted)    & 80.52 & 81.13 & 79.24 & 36.06 & 25.03 & 17.47 & 70.69 & 64.42 & 55.08 \\
\midrule
FixMatch$^\dagger$~\cite{sohn2020fixmatch}       & 64.45 & 60.67 & 61.33 & 33.45 & 11.78 & 10.74 & 30.23 & 17.40 &  9.78 \\
FlexMatch$^\dagger$~\cite{zhang2021flexmatch}     & 69.23 & 68.45 & 67.89 & 48.41 & 34.03 & 42.13 & 35.78 & 32.82 & 18.71 \\
SPICE$^\dagger$~\cite{niu2022spice}               & 83.73 & 80.89 & 82.40 & 51.62 & 34.45 & 38.67 & 48.71 & 46.40 & 32.45 \\
FixMatch~\cite{sohn2020fixmatch}       & 85.32 & 79.93 & 75.17 & 68.26 & 31.23 & 38.17 & 71.78 & 62.11 & 41.70 \\
FlexMatch~\cite{zhang2021flexmatch}    & 87.13 & 80.24 & 73.50 & 69.17 & 30.43 & 36.51 & 72.84 & 64.32 & 38.89 \\
SPICE~\cite{niu2022spice}              & 94.10 & 89.11 & 91.09 & 53.83 & 21.52 & 33.04 & 74.78 & 60.03 & 40.67 \\
w/o $k$-NN Consistency       & 95.23 & 90.04 & 91.01 & 68.78 & 28.45 & 36.43 & 74.31 & 57.17 & 33.90 \\
w/o Confidence Selection      & 95.82 & 90.34 & 91.53 & 67.70 & 34.22 & 35.34 & 73.92 & 60.04 & 39.63 \\
\midrule
w/o Topology branch    & 94.91 & 90.18 & 91.67 & 67.23 & 30.78 & 36.40 & 74.82 & 52.73 & 40.08 \\
Fix $\lambda_{\text{topo}}\!=\!0.5$    & 91.72 & 90.78 & 90.63 & 65.76 & 30.43 & 28.41 & 71.50 & 52.21 & 40.13 \\
Fix $\lambda_{\text{topo}}\!=\!0.2$    & 95.80 & \textbf{92.93} & 93.04 & 68.72 & 34.45 & 34.51 & 75.18 & 58.67 & 39.23 \\
w/o Phase A            & 81.62 & 83.53 & 74.17 & 63.56 & 27.42 & 30.78 & 65.45 & 52.20 & 32.01 \\
w/o Phase B            & 95.13 & 88.56 & 87.33 & 68.16 & 41.04 & 38.30 & 75.62 & 60.28 & 39.31 \\
\midrule
\textbf{VTRL-Med (Full)} & \textbf{96.22} & 92.64 & \textbf{93.77} & \textbf{71.51} & \textbf{47.90} & \textbf{39.43} & \textbf{77.17} & \textbf{63.31} & \textbf{43.35} \\
\bottomrule
\end{tabular}
}

\end{table}

\rev{We systematically evaluate each component of VTRL-Med on three datasets (Table~\ref{tab:ablation_study}). Comparisons of SSL methods and SAM variants are in Supplementary.}

\noindent\textbf{Stage Necessity \rev{and Their Complementary Roles}.} Removing Stage 1 (SAM Adaptation) causes catastrophic failures (ACC drops by 30--39\%)\rev{, confirming that structural priors provide the essential discriminative foundation}. Removing Stage 2 (Pseudo-Labeling) incurs dataset-dependent effects: small drops on PathMNIST ($-3.6$\%) but larger drops on ISIC2019 ($-16.5$\%) and Kvasir ($-7.9$\%), indicating that Stage 2 becomes more critical for complex datasets. \rev{Removing Stage 3 leads to substantial degradation (9--32\%), and comparing w/o Stage 3 (row 3, 63.69\%) with w/o Topology branch (row 16, 94.91\%) on PathMNIST reveals that Stage 3's visual refinement contributes +31.2\%, while the topology branch adds a further +1.3\%, correcting 25.7\% of the remaining errors on PathMNIST.} \rev{The gap between w/o Stage~3 (row~3, 63.69\%) and SAM adapted with $K$-means (row~7, 80.52\%) on PathMNIST reflects a known limitation of pseudo-label bootstrapping: Stage~2 trains ClusHead from scratch on noisy pseudo-labels with a frozen encoder, so without Stage~3's joint encoder fine-tuning, ClusHead overfits to early pseudo-label errors that $K$-means on the original features avoids. Stage~3 resolves this by unfreezing the encoder and progressively correcting both the features and the head, yielding 96.22\%.}

\noindent\textbf{Stage 1 Design.} To isolate the effect of the feature backbone, we apply $K$-means directly without any pipeline training (rows 4--7). Raw pixels (23.35--45.78\% ACC) and ImageNet-pretrained ResNet-50 (22.66--73.82\%) confirm that generic representations are insufficient. Frozen SAM yields 25.13--49.51\%, underperforming ResNet-50 on PathMNIST and Kvasir (47.90 vs.\ 73.82\%, 49.51 vs.\ 55.28\%) though comparable on ISIC2019, indicating that SAM's mask-oriented pretraining does not consistently produce discriminative clustering features. Stage 1 adaptation substantially improves SAM to 36.06--80.52\%, reaching performance comparable to full SSL baselines in Table~\ref{tab:main_results} (e.g., 80.52\% vs.\ SimCLR 77.21\%, MoCo-v3 77.53\% on PathMNIST), demonstrating that domain-specific self-distillation, not the pretrained encoder alone, provides the discriminative foundation.

\noindent\textbf{Stage 2 Design.} \rev{We validate our two-step filtering through three comparisons (rows 8--15): (i) applying FixMatch, FlexMatch, and SPICE directly on adapted SAM features without Stage 3 refinement ($^\dagger$, rows 8--10) to test whether simple pseudo-labeling suffices; (ii) replacing our filtering with these methods while keeping Stage 3 unchanged (rows 11--13) to isolate pseudo-label quality; (iii) ablating each step within our filtering (rows 14--15). The $^\dagger$ rows achieve 64.45--83.73\% on PathMNIST, 33.45--51.62\% on ISIC2019, and 30.23--48.71\% on Kvasir, all substantially below VTRL-Med (96.22/71.51/77.17\%). Confidence-based methods (FixMatch with $\tau{=}0.95$; FlexMatch with base threshold $0.95$ and per-class adaptive thresholds) generally fall below SAM + $K$-means (e.g., 64.45\% and 69.23\% vs.\ 80.52\% on PathMNIST; 30.23\% and 35.78\% vs.\ 70.69\% on Kvasir), because the randomly initialized head lacks calibrated confidence for threshold-based filtering (detailed analysis in Supplementary). Adding Stage 3 refinement dramatically improves all methods (e.g., FixMatch: 30.23$\to$71.78\% on Kvasir), confirming that progressive refinement, not pseudo-label strategy alone, drives performance.} With Stage 3 refinement, our two-step filtering achieves the highest ACC on all three datasets, outperforming FixMatch by 3.3--10.9\%, FlexMatch by 2.3--9.1\%, and SPICE by 2.1--17.7\%. Removing confidence selection drops ACC by 0.4--3.8\% and removing $k$-NN consistency drops ACC by 1.0--2.9\%\rev{, with all thresholds confirmed robust within $\pm$1.5\% (see Supplementary)}.

\noindent\textbf{Stage 3 Design.} \rev{The topology branch (row 16) improves ACC by 1.3--4.3\% across datasets.} Fixed high weight ($\lambda_{\text{topo}}{=}0.5$) substantially harms performance (4.5--5.8\% drops), while fixed moderate weight ($\lambda_{\text{topo}}{=}0.2$) remains competitive (0.4--2.8\% drops)\rev{, justifying the phase-dependent weighting ($\lambda_{\text{topo}}^A{=}0.5$ for bootstrap, $\lambda_{\text{topo}}^B{=}0.2$ for refinement)}. Crucially, removing Phase A (Bootstrap TopoNet) causes severe degradation (8.0--14.6\% drops), indicating that the topology branch requires direct supervision before mutual \rev{refinement}. Removing Phase B yields moderate drops (1.1--3.4\%).

\begin{figure*}[t!]
\centering
\includegraphics[width=0.85\linewidth]{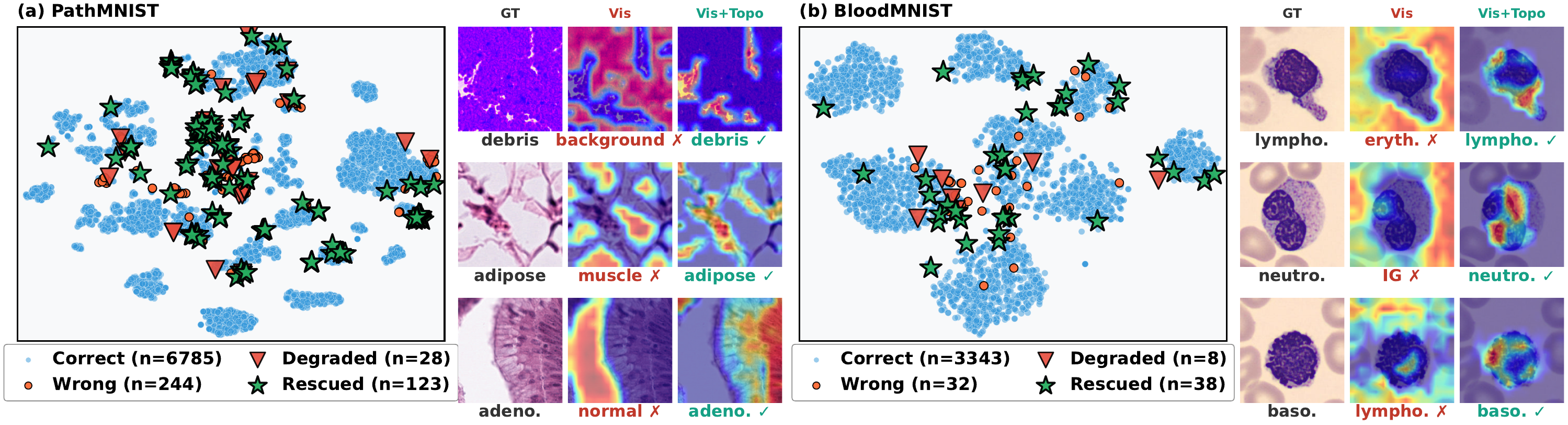}
\caption{\rev{t-SNE visualizations of embeddings for (a) PathMNIST and (b) BloodMNIST. Samples are color-coded: Correct (blue) = both branches predict correctly, Wrong (orange) = both predict incorrectly, Rescued (green stars) = topology branch corrects the visual branch, Degraded (red triangles) = topology branch overrides a correct visual prediction. The right columns show Grad-CAM attention of the visual branch alone (Vis) vs.\ with topology guidance (Vis+Topo) for representative rescued cases.}}
\label{fig:topology_boundary}

\end{figure*}

\subsection{Analysis of Topology-Corrected Predictions}
\label{sec:boundary_analysis}

\rev{Section~\ref{sec:ablation} shows that the topology branch improves ACC by 1.3--4.3\% across datasets.}
To understand \emph{where} and \emph{why} this occurs, we analyze the positions of topology-corrected samples in the feature space and attention patterns (Fig.~\ref{fig:topology_boundary}).

\noindent\textbf{Where Topology Helps.}
We visualize all test samples with t-SNE~\cite{maaten2008visualizing} \rev{(Fig.~\ref{fig:topology_boundary})} and quantify boundary proximity using $k$-NN impurity ($k{=}10$), defined as the proportion of neighbors belonging to different clusters. 
\rev{The rescued} samples (green stars) concentrate near cluster boundaries with significantly higher $k$-NN impurity than correctly clustered samples ($p{<}0.001$), confirming that they lie in ambiguous regions\rev{: the rescued samples are 4.3$\times$ and 10.3$\times$ closer to class boundaries than correctly classified samples on PathMNIST and BloodMNIST, respectively}.
\rev{The net correction is consistently positive: on PathMNIST, topology rescues 123 samples while degrading 28 (net +95, 4.4$\times$ rescue-to-degrade ratio); on BloodMNIST, it rescues 38 while degrading 8 (net +30, 4.75$\times$ ratio).}
\rev{Degraded cases (red triangles) also concentrate near cluster boundaries, confirming that topology acts as a boundary-sensitive regularizer that occasionally over-corrects. Per-dataset counts across all 13 benchmarks and per-class confusion analysis are in Supplementary.}

\noindent\textbf{Why Topology Helps.}
We apply Grad-CAM~\cite{selvaraju2017grad} to representative \rev{rescued} cases (Fig.~\ref{fig:topology_boundary}, right columns), and compare the attention patterns of the visual branch trained \emph{without} \rev{(Vis)} vs.\ \emph{with} topology guidance \rev{(Vis+Topo)}. 
Topology-guided training consistently pushes attention toward more discriminative structures: on PathMNIST, it broadens focus to tissue-level organization beyond localized regions; on BloodMNIST, it concentrates on diagnostic cell morphology rather than dispersing across cells and background. This explains why topology rescues mostly the boundary cases that visual features alone cannot resolve.

\begin{table}[bt]
\centering
\caption{Computational costs and clustering accuracy (avg. over 13 datasets), measured on TissueMNIST, with 300 epochs and batch size 256. Tput.\,=\,Throughput.}
\label{tab:efficiency}
\scriptsize  
\setlength{\tabcolsep}{4pt}  
\begin{tabular}{@{}lrrrrr@{}}  
\toprule
\textbf{Method} & \textbf{Train} & \textbf{Params} & \textbf{FLOPs} & \textbf{Tput.} & \textbf{ACC} \\
& \textbf{(h)$\downarrow$} & \textbf{(M)$\downarrow$} & \textbf{(G)$\downarrow$} & \textbf{(img/s)$\uparrow$} & \textbf{(\%)$\uparrow$} \\
\midrule
DINOv3-B~\cite{simeoni2025dinov3} & 33.41 & 85.80 & 17.58 & 3522.33 & 60.88 \\
SPICE~\cite{niu2022spice} & 31.41 & 46.71 & 7.84 & 4090.40 & 77.03 \\
CDC~\cite{jia2024towards} & 40.28 & 24.56 & \textbf{4.11} & \textbf{6217.12} & 70.33 \\
MedTrans~\cite{jahanian2024medtranscluster} & \textbf{26.82} & 94.51 & 4.61 & 5252.68 & 65.10 \\
DeepMCAT~\cite{kart2021deepmcat} & 35.43 & \textbf{14.75} & 15.37 & 3487.67 & 67.93 \\
MedMamba-B~\cite{yue2024medmamba} & 36.72 & 46.87 & 7.31 & 876.92 & \textbf{84.80} \\
\midrule
VTRL-Med (Ours) & 36.87 & 90.30 & 17.54 & 3859.00 & 81.51 \\
\bottomrule
\end{tabular}

\end{table}

\begin{table}[bt]
\centering
\caption{VTRL-Med cost breakdown. (a) Training (300 epochs). (b) Inference. Both measured on TissueMNIST, with batch size 256. Tput. = Throughput; PH = Persistent Homology.}
\label{tab:component_cost}
\scriptsize
\setlength{\tabcolsep}{2pt}
\begin{minipage}[t]{0.48\columnwidth}  
\centering
\subcaption{Training cost breakdown}
\label{tab:cost_train}
\vspace{0.1cm}
\begin{tabular}[t]{lrr}
\toprule
\textbf{Component} & \textbf{Time (h)} & \textbf{\%} \\
\midrule
SAM Self-Distill & 16.20 & 44.0 \\
Gram Anchor & 5.70 & 15.5 \\
Hi-Res Refine & 4.76 & 12.9 \\
\midrule
Feature Extract & 4.23 & 11.5 \\
Prototype Filter & 1.63 & 4.4 \\
Head Training & 0.02 & 0.1 \\
\midrule
Reliable Select & 0.03 & 0.1 \\
Visual Branch & 4.26 & 11.6 \\
Topology Branch & 0.04 & 0.1 \\
\midrule
\textbf{Total} & \textbf{36.87} & \textbf{100.0} \\
\bottomrule
\end{tabular}
\end{minipage}%
\hspace{0.04\columnwidth}
\begin{minipage}[t]{0.48\columnwidth}  
\centering
\subcaption{Inference cost breakdown}
\label{tab:cost_infer}
\vspace{0.1cm}
\begin{tabular}[t]{lrrr}
\toprule
\textbf{Component} & \textbf{Params} & \textbf{FLOPs} & \textbf{Tput.} \\
& \textbf{(M)} & \textbf{(G)} & \textbf{(img/s)} \\
\midrule
\textit{Vis. Branch} & 90.1 & 17.5 & 3895 \\
\quad SAM & 89.7 & 17.5 & 3896 \\
\quad ClusHead & 0.4 & 0.01 & 4838k \\
\midrule
\textit{Topo. Branch} & 0.2 & 0.02 & 376k \\
\midrule
\textbf{Full Model} & 90.3 & 17.5 & 3859 \\
\textit{w/o Topology} & 90.1 & 17.5 & 3895 \\
\bottomrule
\end{tabular}
\end{minipage}

\end{table}

\subsection{Computational Complexity}
\label{sec:cost}

Table~\ref{tab:efficiency} and Table~\ref{tab:component_cost} report computational costs, measured on TissueMNIST (NVIDIA H100, batch size 256). VTRL-Med's training time (36.87h) is competitive with Transformer-based methods; its model size (90.30M parameters) and throughput (3859 img/s) are comparable to other large-scale approaches. Notably, VTRL-Med attains 4.4$\times$ higher throughput than the supervised MedMamba-B (876.92 img/s) with only 3.29\% lower accuracy, giving favorable accuracy-efficiency trade-offs for \rev{label-free} clustering.
Table~\ref{tab:component_cost} shows Stage 1 dominating the training time (72.4\%), especially SAM self-distillation (44.0\%); Stages 2 \& 3 contribute 15.9\% and 11.7\%. PH feature extraction \rev{(46.8 min)} is offline. For inference, SAM takes 99.3\% of parameters and most FLOPs, while topology adds minimal overhead (0.2M parameters, 0.02G FLOPs, $<$1\% throughput overhead). \rev{This confirms that topological regularization improves clustering with small computation cost.}

\section{Conclusions}
\label{sec:conclusion}

We presented VTRL-Med, a label-free medical image clustering framework that unifies SAM-based structural priors with persistent-homology-based topological regularization. Experiments on 13 benchmarks confirmed their complementary roles, with VTRL-Med achieving SOTA label-free performance within 3.29\% of supervised accuracy at negligible additional cost. This structural-and-topological perspective opens new avenues for anatomy-faithful deep medical image clustering.

{
    \small
    \bibliographystyle{ieeenat_fullname}
    \bibliography{main}

@String(CVPR= {IEEE Conf. Comput. Vis. Pattern Recog.})

@String(ECCV= {Eur. Conf. Comput. Vis.})

@String(AAAI = {AAAI})

@String(CVPR  = {CVPR})

@String(ECCV  = {ECCV})

@article{litjens2017survey,
  title={A survey on deep learning in medical image analysis},
  author={Litjens, Geert and Kooi, Thijs and Bejnordi, Babak Ehteshami and Setio, Arnaud Arindra Adiyoso and Ciompi, Francesco and Ghafoorian, Mohsen and Van Der Laak, Jeroen Awm and Van Ginneken, Bram and S{\'a}nchez, Clara I},
  journal={Medical Image Analysis},
  volume={42},
  pages={60--88},
  year={2017},
  publisher={Elsevier}
}

@inproceedings{irvin2019chexpert,
  title={{CheXpert}: A large chest radiograph dataset with uncertainty labels and expert comparison},
  author={Irvin, Jeremy and Rajpurkar, Pranav and Ko, Michael and Yu, Yifan and Ciurea-Ilcus, Silviana and Chute, Chris and Marklund, Henrik and Haghgoo, Behzad and Ball, Robyn and Shpanskaya, Katie and others},
  booktitle={Proceedings of the AAAI Conference on Artificial Intelligence},
  volume={33},
  pages={590--597},
  year={2019}
}

@article{ren2024deep,
  title={Deep clustering: A comprehensive survey},
  author={Ren, Yazhou and Pu, Jingyu and Yang, Zhimeng and Xu, Jie and Li, Guofeng and Pu, Xiaorong and Yu, Philip S and He, Lifang},
  journal={IEEE Transactions on Neural Networks and Learning Systems},
  volume={36},
  number={4},
  pages={5858--5878},
  year={2024},
  publisher={IEEE}
}

@article{zhou2024comprehensive,
  title={A comprehensive survey on deep clustering: Taxonomy, challenges, and future directions},
  author={Zhou, Sheng and Xu, Hongjia and Zheng, Zhuonan and Chen, Jiawei and Li, Zhao and Bu, Jiajun and Wu, Jia and Wang, Xin and Zhu, Wenwu and Ester, Martin},
  journal={ACM Computing Surveys},
  volume={57},
  number={3},
  pages={1--38},
  year={2024},
  publisher={ACM New York, NY}
}

@article{seebock2018unsupervised,
  title={Unsupervised identification of disease marker candidates in retinal {OCT} imaging data},
  author={Seeb{\"o}ck, Philipp and Waldstein, Sebastian M and Klimscha, Sophie and Bogunovic, Hrvoje and Schlegl, Thomas and Gerendas, Bianca S and Donner, Rene and Schmidt-Erfurth, Ursula and Langs, Georg},
  journal={IEEE Transactions on Medical Imaging},
  volume={38},
  number={4},
  pages={1037--1047},
  year={2018},
  publisher={IEEE}
}

@inproceedings{xie2016unsupervised,
  title={Unsupervised deep embedding for clustering analysis},
  author={Xie, Junyuan and Girshick, Ross and Farhadi, Ali},
  booktitle={Proceedings of the International Conference on Machine Learning},
  pages={478--487},
  year={2016}
}

@inproceedings{van2020scan,
  title={{SCAN}: Learning to classify images without labels},
  author={Van Gansbeke, Wouter and Vandenhende, Simon and Georgoulis, Stamatios and Proesmans, Marc and Van Gool, Luc},
  booktitle={Proceedings of the European Conference on Computer Vision},
  pages={268--285},
  year={2020}
}

@article{niu2022spice,
  title={{SPICE}: Semantic pseudo-labeling for image clustering},
  author={Niu, Chuang and Shan, Hongming and Wang, Ge},
  journal={IEEE Transactions on Image Processing},
  volume={31},
  pages={7264--7278},
  year={2022},
  publisher={IEEE}
}

@inproceedings{chen2020simple,
  title={A simple framework for contrastive learning of visual representations},
  author={Chen, Ting and Kornblith, Simon and Norouzi, Mohammad and Hinton, Geoffrey},
  booktitle={Proceedings of the International Conference on Machine Learning},
  pages={1597--1607},
  year={2020}
}

@article{rajpurkar2017chexnet,
  title={{CheXNet}: Radiologist-level pneumonia detection on chest {X-Rays} with deep learning},
  author={Rajpurkar, Pranav and Irvin, Jeremy and Zhu, Kaylie and Yang, Brandon and Mehta, Hershel and Duan, Tony and Ding, Daisy and Bagul, Aarti and Langlotz, Curtis and Shpanskaya, Katie and others},
  journal={arXiv preprint arXiv:1711.05225},
  year={2017}
}

@article{campanella2019clinical,
  title={Clinical-grade computational pathology using weakly supervised deep learning on whole slide images},
  author={Campanella, Gabriele and Hanna, Matthew G and Geneslaw, Luke and Miraflor, Allen and Werneck Krauss Silva, Vitor and Busam, Klaus J and Brogi, Edi and Reuter, Victor E and Klimstra, David S and Fuchs, Thomas J},
  journal={Nature Medicine},
  volume={25},
  number={8},
  pages={1301--1309},
  year={2019},
  publisher={Nature Publishing Group US New York}
}

@article{aljalbout2018clustering,
  title={Clustering with deep learning: Taxonomy and new methods},
  author={Aljalbout, Elie and Golkov, Vladimir and Siddiqui, Yawar and Strobel, Maximilian and Cremers, Daniel},
  journal={arXiv preprint arXiv:1801.07648},
  year={2018}
}

@inproceedings{asano2020self,
  title={Self-labelling via simultaneous clustering and representation learning},
  author={Asano, Yuki M and Rupprecht, Christian and Vedaldi, Andrea},
  booktitle={Proceedings of the International Conference on Learning Representations},
  year={2020}
}

@inproceedings{caron2020unsupervised,
  title={Unsupervised learning of visual features by contrasting cluster assignments},
  author={Caron, Mathilde and Misra, Ishan and Mairal, Julien and Goyal, Priya and Bojanowski, Piotr and Joulin, Armand},
  booktitle={Proceedings of the Conference on Neural Information Processing Systems},
  volume={33},
  pages={9912--9924},
  year={2020}
}

@article{min2018survey,
  title={A survey of clustering with deep learning: From the perspective of network architecture},
  author={Min, Erxue and Guo, Xifeng and Liu, Qiang and Zhang, Gen and Cui, Jianjing and Long, Jun},
  journal={IEEE Access},
  volume={6},
  pages={39501--39514},
  year={2018}
}

@article{medmnistv2,
    title={{MedMNIST v2}-A large-scale lightweight benchmark for {2D and 3D} biomedical image classification},
    author={Yang, Jiancheng and Shi, Rui and Wei, Donglai and Liu, Zequan and Zhao, Lin and Ke, Bilian and Pfister, Hanspeter and Ni, Bingbing},
    journal={Scientific Data},
    volume={10},
    number={1},
    pages={41},
    year={2023},
    publisher={Nature Publishing Group UK London}
}

@inproceedings{medmnistv1,
    title={{MedMNIST} Classification Decathlon: A Lightweight AutoML Benchmark for Medical Image Analysis},
    author={Yang, Jiancheng and Shi, Rui and Ni, Bingbing},
    booktitle={Proceedings of the IEEE 18th International Symposium on Biomedical Imaging},
    pages={191--195},
    year={2021}
}

@article{tschandl2018ham10000,
  title={The {HAM10000} dataset, a large collection of multi-source dermatoscopic images of common pigmented skin lesions},
  author={Tschandl, Philipp and Rosendahl, Cliff and Kittler, Harald},
  journal={Scientific Data},
  volume={5},
  pages={180161},
  year={2018},
  publisher={Nature Publishing Group}
}

@inproceedings{pogorelov2017kvasir,
  title={Kvasir: A multi-class image dataset for computer aided gastrointestinal disease detection},
  author={Pogorelov, Konstantin and Randel, Kristin Ranheim and Griwodz, Carsten and Eskeland, Sigrun Losada and de Lange, Thomas and Johansen, Dag and Spampinato, Concetto and Dang-Nguyen, Duc-Tien and Lux, Mathias and Schmidt, Peter Thelin and others},
  booktitle={Proceedings of the 8th ACM on Multimedia Systems Conference},
  pages={164--169},
  year={2017}
}

@article{jahanian2024medtranscluster,
  title={{MedTransCluster}: Transfer learning for deep medical image clustering},
  author={Jahanian, Mojtaba and Karimi, Abbas and Eraghi, Nafiseh Osati and Zarafshan, Faraneh},
  journal={Intelligence-Based Medicine},
  volume={9},
  pages={100139},
  year={2024},
  publisher={Elsevier}
}

@article{jia2024towards,
  title={Towards calibrated deep clustering network},
  author={Jia, Yuheng and Cheng, Jianhong and Liu, Hui and Hou, Junhui},
  journal={arXiv preprint arXiv:2403.02998},
  year={2024}
}

@inproceedings{kart2021deepmcat,
  title={{DeepMCAT}: Large-scale deep clustering for medical image categorization},
  author={Kart, Turkay and Bai, Wenjia and Glocker, Ben and Rueckert, Daniel},
  booktitle={Proceedings of the MICCAI Workshop on Deep Generative Models},
  pages={259--267},
  year={2021}
}

@article{kuhn1955hungarian,
  title={The {Hungarian} method for the assignment problem},
  author={Kuhn, Harold W},
  journal={Naval Research Logistics Quarterly},
  volume={2},
  number={1-2},
  pages={83--97},
  year={1955}
}

@article{strehl2002cluster,
  title={Cluster ensembles---A knowledge reuse framework for combining multiple partitions},
  author={Strehl, Alexander and Ghosh, Joydeep},
  journal={Journal of Machine Learning Research},
  volume={3},
  pages={583--617},
  year={2002}
}

@article{hubert1985comparing,
  title={Comparing partitions},
  author={Hubert, Lawrence and Arabie, Phipps},
  journal={Journal of Classification},
  volume={2},
  pages={193--218},
  year={1985}
}

@inproceedings{chen2021mocov3,
  title={An empirical study of training self-supervised vision {Transformers}},
  author={Chen, Xinlei and Xie, Saining and He, Kaiming},
  booktitle={Proceedings of the IEEE/CVF International Conference on Computer Vision},
  pages={9640--9649},
  year={2021}
}

@inproceedings{grill2020bootstrap,
  title={{Bootstrap Your Own Latent}: A new approach to self-supervised learning},
  author={Grill, Jean-Bastien and Strub, Florian and Altch{\'e}, Florent and Tallec, Corentin and Richemond, Pierre H and Buchatskaya, Elena and Doersch, Carl and Pires, Bernardo Avila and Guo, Zhaohan Daniel and Azar, Mohammad Gheshlaghi and others},
  booktitle={Proceedings of the Conference on Neural Information Processing Systems},
  volume={33},
  pages={21271--21284},
  year={2020}
}

@article{simeoni2025dinov3,
  title={{DINOv3}},
  author={Sim{\'e}oni, Oriane and Vo, Huy V and Seitzer, Maximilian and Baldassarre, Federico and Oquab, Maxime and Jose, Cijo and Khalidov, Vasil and Szafraniec, Marc and Yi, Seungeun and Ramamonjisoa, Micha{\"e}l and others},
  journal={arXiv preprint arXiv:2508.10104},
  year={2025}
}

@article{yue2024medmamba,
  title={{MedMamba}: Vision {Mamba} for medical image classification},
  author={Yue, Yubiao and Li, Zhenzhang},
  journal={arXiv preprint arXiv:2403.03849},
  year={2024}
}

@inproceedings{kirillov2023segment,
  title={Segment anything},
  author={Kirillov, Alexander and Mintun, Eric and Ravi, Nikhila and Mao, Hanzi and Rolland, Chloe and Gustafson, Laura and Xiao, Tete and Whitehead, Spencer and Berg, Alexander C and Lo, Wan-Yen and others},
  booktitle={Proceedings of the IEEE/CVF International Conference on Computer Vision},
  pages={4015--4026},
  year={2023}
}

@book{edelsbrunner2010computational,
  title={Computational Topology: An Introduction},
  author={Edelsbrunner, Herbert and Harer, John},
  year={2010},
  publisher={American Mathematical Society}
}

@article{chambers2025stable,
  title={A Stable and Theoretically Grounded {Gromov-Wasserstein} Distance for {Reeb} Graph Comparison using Persistence Images},
  author={Chambers, Erin W and Meng, Guangyu},
  journal={arXiv preprint arXiv:2507.01171},
  year={2025}
}

@inproceedings{cohen2005stability,
  title={Stability of persistence diagrams},
  author={Cohen-Steiner, David and Edelsbrunner, Herbert and Harer, John},
  booktitle={Proceedings of the Twenty-first Annual Symposium on Computational Geometry},
  pages={263--271},
  year={2005}
}

@article{qaiser2019fast,
  title={Fast and accurate tumor segmentation of histology images using persistent homology and deep convolutional features},
  author={Qaiser, Talha and Tsang, Yee-Wah and Taniyama, Daiki and Sakamoto, Naoya and Nakane, Kazuaki and Epstein, David and Rajpoot, Nasir},
  journal={Medical Image Analysis},
  volume={55},
  pages={1--14},
  year={2019},
  publisher={Elsevier}
}

@article{lawson2019persistent,
  title={Persistent homology for the quantitative evaluation of architectural features in prostate cancer histology},
  author={Lawson, Peter and Sholl, Andrew B and Brown, J Quincy and Fasy, Brittany Terese and Wenk, Carola},
  journal={Scientific Reports},
  volume={9},
  number={1},
  pages={1--12},
  year={2019},
  publisher={Nature Publishing Group}
}

@inproceedings{peng2024phg,
  title={{PHG-Net}: Persistent homology guided medical image classification},
  author={Peng, Yaopeng and Wang, Hongxiao and Sonka, Milan and Chen, Danny Z},
  booktitle={Proceedings of the IEEE/CVF Winter Conference on Applications of Computer Vision},
  pages={7583--7592},
  year={2024}
}

@incollection{wagner2011efficient,
  title={Efficient computation of persistent homology for cubical data},
  author={Wagner, Hubert and Chen, Chao and Vu{\c{c}}ini, Erald},
  booktitle={Proceedings of the Topological Methods in Data Analysis and Visualization II: Theory, Algorithms, and Applications},
  pages={91--106},
  year={2011},
  publisher={Springer}
}

@inproceedings{singh2014topological,
  title={Topological Descriptors of Histology Images},
  author={Singh, Nikhil and Couture, Heather D and Marron, J. S. and Perou, Charles and Niethammer, Marc},
  booktitle={Proceedings of the Machine Learning in Medical Imaging},
  pages={231--239},
  year={2014},
  volume={8679}
}

@article{elston1991pathological,
  title={Pathological prognostic factors in breast cancer. {I}. The value of histological grade in breast cancer: Experience from a large study with long-term follow-up},
  author={Elston, Christopher W and Ellis, Ian O},
  journal={Histopathology},
  volume={19},
  number={5},
  pages={403--410},
  year={1991},
  publisher={Wiley Online Library}
}

@article{sohn2020fixmatch,
  title={{FixMatch}: Simplifying semi-supervised learning with consistency and confidence},
  author={Sohn, Kihyuk and Berthelot, David and Carlini, Nicholas and Zhang, Zizhao and Zhang, Han and Raffel, Colin A and Cubuk, Ekin Dogus and Kurakin, Alexey and Li, Chun-Liang},
  journal={Advances in Neural Information Processing Systems},
  volume={33},
  pages={596--608},
  year={2020}
}

@article{maaten2008visualizing,
  title={Visualizing data using {t-SNE}},
  author={Maaten, Laurens van der and Hinton, Geoffrey},
  journal={Journal of Machine Learning Research},
  volume={9},
  number={Nov},
  pages={2579--2605},
  year={2008}
}

@inproceedings{selvaraju2017grad,
  title={{Grad-CAM}: Visual explanations from deep networks via gradient-based localization},
  author={Selvaraju, Ramprasaath R and Cogswell, Michael and Das, Abhishek and Vedantam, Ramakrishna and Parikh, Devi and Batra, Dhruv},
  booktitle={Proceedings of the IEEE International Conference on Computer Vision},
  pages={618--626},
  year={2017}
}

@inproceedings{SAM2,
title={{SAM} 2: Segment Anything in Images and Videos},
    author={Nikhila Ravi and Valentin Gabeur and Yuan-Ting Hu and Ronghang Hu and Chaitanya Ryali and Tengyu Ma and Haitham Khedr and Roman R{\"a}dle and Chloe Rolland and Laura Gustafson and Eric Mintun and Junting Pan and Kalyan Vasudev Alwala and Nicolas Carion and Chao-Yuan Wu and Ross Girshick and Piotr Dollar and Christoph Feichtenhofer},
    booktitle={Proceedings of the International Conference on Learning Representations},
    year={2025}
}

@article{MedSAM2,
    title={{MedSAM2}: Segment Anything in {3D} Medical Images and Videos},
    author={Ma, Jun and Yang, Zongxin and Kim, Sumin and Chen, Bihui and Baharoon, Mohammed and Fallahpour, Adibvafa and Asakereh, Reza and Lyu, Hongwei and Wang, Bo},
    journal={arXiv preprint arXiv:2504.03600},
    year={2025}
}

@inproceedings{caron2021emerging,
  title={Emerging properties in self-supervised {Vision Transformers}},
  author={Caron, Mathilde and Touvron, Hugo and Misra, Ishan and J{\'e}gou, Herv{\'e} and Mairal, Julien and Bojanowski, Piotr and Joulin, Armand},
  booktitle={Proceedings of the IEEE/CVF International Conference on Computer Vision},
  pages={9650--9660},
  year={2021}
}

@InProceedings{Azizi_2021_ICCV,
    author    = {Azizi, Shekoofeh and Mustafa, Basil and Ryan, Fiona and Beaver, Zachary and Freyberg, Jan and Deaton, Jonathan and Loh, Aaron and Karthikesalingam, Alan and Kornblith, Simon and Chen, Ting and Natarajan, Vivek and Norouzi, Mohammad},
    title     = {Big Self-Supervised Models Advance Medical Image Classification},
    booktitle = {Proceedings of the IEEE/CVF International Conference on Computer Vision},
    year      = {2021},
    pages     = {3478-3488}
}

@article{oquab2023dinov2,
  title={{DINOv2}: Learning robust visual features without supervision},
  author={Oquab, Maxime and Darcet, Timoth{\'e}e and Moutakanni, Th{\'e}o and Vo, Huy and Szafraniec, Marc and Khalidov, Vasil and Fernandez, Pierre and Haziza, Daniel and Massa, Francisco and El-Nouby, Alaaeldin and others},
  journal={arXiv preprint arXiv:2304.07193},
  year={2023}
}

@article{MedSAM,
  title={Segment Anything in Medical Images},
  author={Ma, Jun and He, Yuting and Li, Feifei and Han, Lin and You, Chenyu and Wang, Bo},
  journal={Nature Communications},
  volume={15},
  pages={654},
  year={2024}
}

@article{wang2004ssim,
  title={Image quality assessment: From error visibility to structural similarity},
  author={Wang, Zhou and Bovik, Alan C and Sheikh, Hamid R and Simoncelli, Eero P},
  journal={IEEE Transactions on Image Processing},
  volume={13},
  number={4},
  pages={600--612},
  year={2004},
  publisher={IEEE}
}

@article{wu2017unsupervised,
  title={Unsupervised clustering of quantitative image phenotypes reveals breast cancer subtypes with distinct prognoses and molecular pathways},
  author={Wu, Jia and Cui, Yi and Sun, Xiaoli and Cao, Guohong and Li, Bailiang and Ikeda, Debra M and Kurian, Allison W and Li, Ruijiang},
  journal={Clinical Cancer Research},
  volume={23},
  number={13},
  pages={3334--3342},
  year={2017},
  publisher={AACR}
}

@inproceedings{muhammad2019unsupervised,
  title={Unsupervised subtyping of cholangiocarcinoma using a deep clustering convolutional autoencoder},
  author={Muhammad, Hassan and Sigel, Carlie S and Campanella, Gabriele and Boesmueller, Hans and Pak, Linda M and B{\"u}ttner, Reinhard and IJzermans, Jan NM and Jarnagin, William R and Simpson, Amber L and Fuchs, Thomas J},
  booktitle={Proceedings of the International Conference on Medical Image Computing and Computer-Assisted Intervention},
  pages={604--612},
  year={2019}
}

@inproceedings{deng2009imagenet,
  title={{ImageNet}: A large-scale hierarchical image database},
  author={Deng, Jia and Dong, Wei and Socher, Richard and Li, Li-Jia and Li, Kai and Fei-Fei, Li},
  booktitle={2009 IEEE Conference on Computer Vision and Pattern Recognition},
  pages={248--255},
  year={2009}
}

@article{zhang2021flexmatch,
  title={{FlexMatch}: Boosting semi-supervised learning with curriculum pseudo labeling},
  author={Zhang, Bowen and Wang, Yidong and Hou, Wenxin and Wu, Hao and Wang, Jindong and Okumura, Manabu and Shinozaki, Takahiro},
  journal={Advances in Neural Information Processing Systems},
  volume={34},
  pages={18408--18419},
  year={2021}
}

@inproceedings{zhu2024advancing,
  title={Advancing ultrasound medical continuous learning with task-specific generalization and adaptability},
  author={Zhu, Chunzheng and Lin, Jianxin and Tan, Guanghua and Zhu, Ningbo and Li, Kenli and Wang, Chunlian and Li, Shengli},
  booktitle={2024 IEEE International Conference on Bioinformatics and Biomedicine (BIBM)},
  pages={3019--3025},
  year={2024},
  organization={IEEE}
}

@article{zhu2025fmri2ges,
  title={fMRI2GES: Co-speech gesture reconstruction from fmri signal with dual brain decoding alignment},
  author={Zhu, Chunzheng and Shao, Jialin and Lin, Jianxin and Wang, Yijun and Wang, Jing and Tang, Jinhui and Li, Kenli},
  journal={IEEE Transactions on Circuits and Systems for Video Technology},
  volume={35},
  number={9},
  pages={9017--9029},
  year={2025},
  publisher={IEEE}
}

@inproceedings{zhu2025pathology,
  title={Pathology-aware prototype evolution via llm-driven semantic disambiguation for multicenter diabetic retinopathy diagnosis},
  author={Zhu, Chunzheng and Lin, Yangfang and Shao, Jialin and Lin, Jianxin and Wang, Yijun},
  booktitle={Proceedings of the 33rd ACM International Conference on Multimedia},
  pages={9196--9205},
  year={2025}
}

@article{yao2025learning,
  title={Learning Flow-Guided Registration for RGB-Event Semantic Segmentation},
  author={Yao, Zhen and Ying, Xiaowen and Zhu, Zhiyu and Chuah, Mooi Choo},
  journal={arXiv preprint arXiv:2505.01548},
  year={2025}
}

@article{yao2026reva,
  title={ReVA: A Scene-Centric Dataset Beyond Repetition for Remote Sensing Video Question Answering},
  author={Yao, Zhen and Wang, Likai and Yang, Yuming and Zheng, Zhihao and Lang, Bo and Tang, Qiuyu and Sheng, Jialu and Xu, Jingqi and Yang, Yuehai and Barker, Jumal and others},
  journal={arXiv preprint arXiv:2609.35507},
  year={2026}
}

@article{yao2026semantic,
  title={From Semantic to Reasoing: A Survey of Language-Guided Segmentation in the Multimodal Era},
  author={Yao, Zhen and Lang, Bo and Zheng, Zhihao and Xu, Haoyan and Yao, Zhengtao and Dong, Zhikang and Chuah, Mooi Choo},
  year={2026},
  publisher={TechRxiv}
}

@inproceedings{gu2025topoimages,
  title={{TopoImages}: Incorporating local topology encoding into deep learning models for medical image classification},
  author={Gu, Pengfei and Wang, Hongxiao and Zhang, Yejia and Li, Huimin and Wang, Chaoli and Chen, Danny},
  booktitle={Proceedings of the 33rd ACM International Conference on Multimedia},
  pages={1938--1947},
  year={2025}
}

@inproceedings{gu2026integrating,
  title={Integrating multi-scale and multi-filtration topological features for medical image classification},
  author={Gu, Pengfei and Li, Huimin and Tang, Haoteng and Xu, Dongkuan and Enriquez, Erik and Kim, DongChul and Fu, Bin and Chen, Danny Z},
  booktitle={Proceedings of the 2026 IEEE/CVF Winter Conference on Applications of Computer Vision},
  pages={8660--8669},
  year={2026}
}

@inproceedings{adame2025topo,
  title={{Topo-VM-UNetV2}: Encoding topology into vision {Mamba UNet} for polyp segmentation},
  author={Adame, Diego and Nunez, Jose A and Vazquez, Fabian and Gurrola, Nayeli and Li, Huimin and Tang, Haoteng and Fu, Bin and Gu, Pengfei},
  booktitle={Proceedings of the 2025 IEEE 38th International Symposium on Computer-Based Medical Systems},
  pages={258--263},
  year={2025}
}

@inproceedings{gu2025self,
  title={Self pre-training with topology-and spatiality-aware masked autoencoders for {3D} medical image segmentation},
  author={Gu, Pengfei and Li, Huimin and Zhang, Yejia and Wang, Chaoli and Chen, Danny Z},
  booktitle={Proceedings of the 2025 IEEE International Conference on Bioinformatics and Biomedicine},
  pages={3608--3613},
  year={2025}
}

@article{meng2025efficient,
  title={Efficient Approximation of {Earth Mover's Distance} Based on Nearest Neighbor Search},
  author={Meng, Guangyu and Zhou, Ruyu and Liu, Liu and Liang, Peixian and Liu, Fang and Chen, Danny Z. and Niemier, Michael and Hu, Xiaobo Sharon},
  journal={IEEE Transactions on Multimedia},
  volume={28},
  pages={227--240},
  year={2026},
  publisher={IEEE},
  doi={10.1109/TMM.2025.3623529}
}

@inproceedings{meng2026topocl,
  title={{TopoCL}: Topological Contrastive Learning for
         Medical Imaging},
  author={Meng, Guangyu and Gu, Pengfei and Liang, Peixian
          and Lalor, John P. and Chambers, Erin Wolf
          and Chen, Danny Z.},
  booktitle={Proceedings of the IEEE/CVF Conference on
             Computer Vision and Pattern Recognition (CVPR)},
  pages={42681--42690},
  year={2026}
}

@inproceedings{meng2026topoagent,
  title={{TopoAgent}: An Agentic Framework for Automated
         Topology Learning in Medical Imaging},
  author={Meng, Guangyu and Gu, Pengfei and Li, Xueyang
          and Shi, Yiyu and Chambers, Erin Wolf
          and Chen, Danny Z.},
  booktitle={Computer Vision -- ECCV 2026},
  series={Lecture Notes in Computer Science},
  volume={17075},
  pages={303--321},
  publisher={Springer},
  year={2026},
  doi={10.1007/978-3-032-37032-7_17}
}

@article{meng2026psychology,
  title={A Psychology-based Unified Dynamic Framework
         for Curriculum Learning},
  author={Meng, Guangyu and Zeng, Qingkai
          and Lalor, John P. and Yu, Hong},
  journal={Computational Linguistics},
  volume={52},
  number={3},
  pages={1001--1049},
  year={2026},
  doi={10.1162/COLI.a.584}
}
}

\end{document}